\documentclass{article}
\usepackage{ijcai18}

\usepackage{times}
\usepackage{xcolor}
\usepackage{soul}
\usepackage[utf8]{inputenc}
\usepackage[small]{caption}

\usepackage{amsfonts,amssymb}
\usepackage{epsfig}
\usepackage{graphicx}
\usepackage{subfigure}
\usepackage{array}
\usepackage{booktabs}

\title{DenseImage Network: Video Spatial-Temporal Evolution \\ Encoding and Understanding}

\author{Xiaokai Chen$^{1,2}$,Ke Gao$^{1}$\\
1 Institute of Computing Technology, Chinese Academy of Sciences, Beijing, China\\
2 University of Chinese Academy of Sciences, Beijing, China\\
\{chenxiaokai,kegao\}@ict.ac.cn}

\begin{document}

\maketitle

\begin{abstract}
Many of the leading approaches for video understanding are data-hungry and time-consuming, failing to capture the gist of spatial-temporal evolution in an efficient manner. The latest research shows that CNN network can reason about static relation of entities in images. To further exploit its capacity in dynamic evolution reasoning, we introduce a novel network module called DenseImage Network(DIN) with two main contributions. 1) A novel compact representation of video which distills its significant spatial-temporal evolution into a matrix called DenseImage, primed for efficient video encoding. 2) A simple yet powerful learning strategy based on DenseImage and a temporal-order-preserving CNN network is proposed for video understanding, which contains a local temporal correlation constraint capturing temporal evolution at multiple time scales with different filter widths. Extensive experiments on two recent challenging benchmarks demonstrate that our DenseImage Network can accurately capture the common spatial-temporal evolution between similar actions, even with enormous visual variations or different time scales. Moreover, we obtain the state-of-the-art results in action and gesture recognition with much less time-and-memory cost, indicating its immense potential in video representing and understanding.
\end{abstract}

\section{Introduction}
The spatial-temporal evolution of entities is of vital importance for video understanding tasks such as action recognition. Similar actions are often performed by various entities at different speeds. Due to the large diversity and complexity, modeling the gist of dynamic evolution in videos is very challenging.

Although deep convolutional neural networks have made significant progress in image understanding tasks~\cite{BadrinarayananK15,VinyalsTBE15}, their impact on video analysis has been somewhat limited, and how to optimally represent videos remains unclear~\cite{SigurdssonRG17}.
\begin{figure}
	\begin{center}
		\includegraphics[width=1\linewidth]{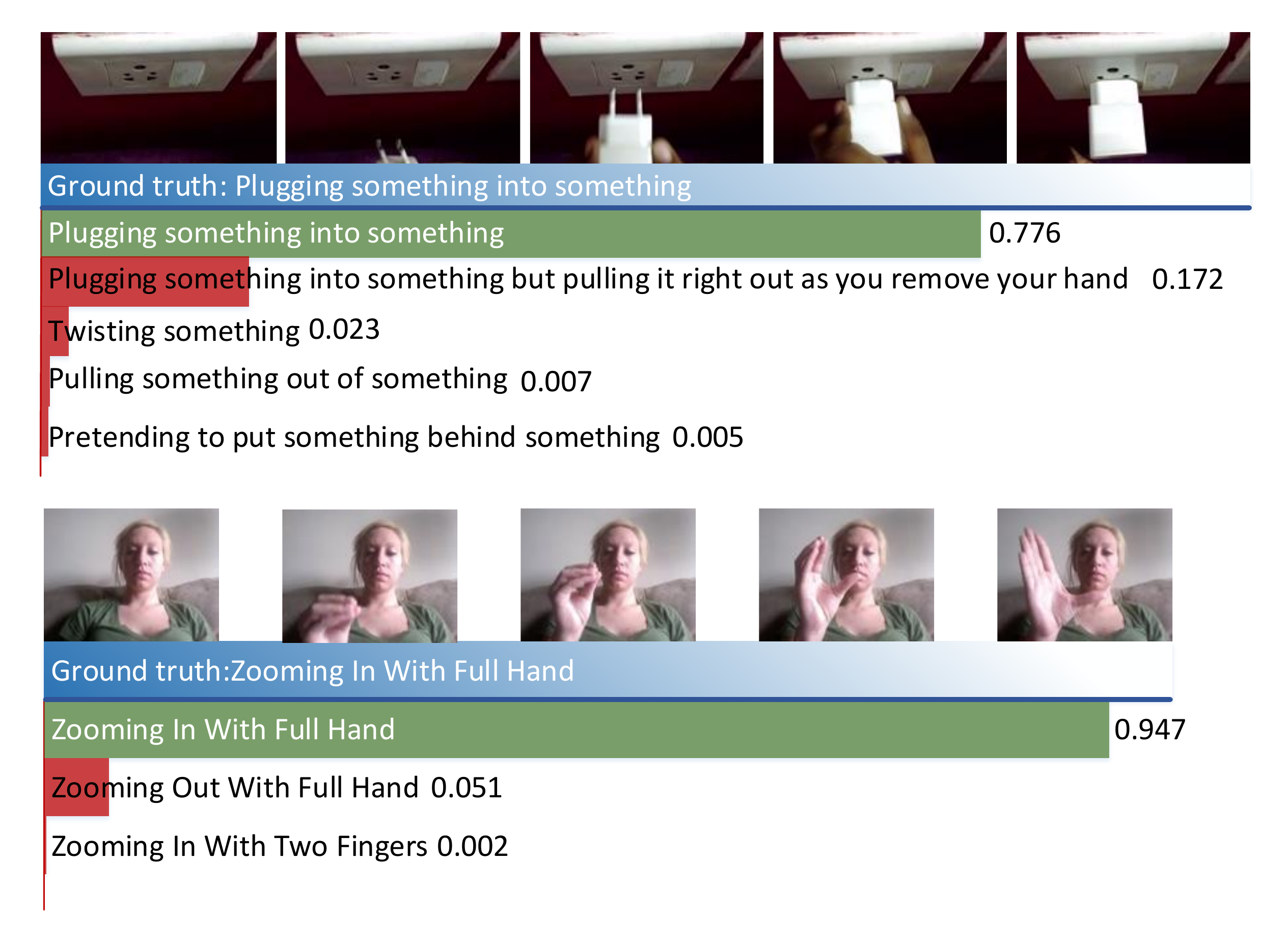}
		\caption{Action recognition results from two challenging new benchmarks Something-Something and Jester. Bars with different lengths indicating the recognition scores.}
		\label{Figure 1.}
	\end{center}
	\vspace{-0.5em}
\end{figure}

On one hand, many top performing approaches~\cite{SimonyanZ14,WangXW0LTG16} rely heavily on optical flow to model temporal dynamics in videos, however, ~\cite{abs-1712-08416} found that the invariance to appearance is responsible for the success of optical flow, rather than the temporal trajectories. 3D CNNs with warm-start are proposed to learn both appearance and motion features simultaneously, but they're data-hungry and computationally expensive based on dense sampled frames~\cite{CarreiraZ17}. Moreover, ~\cite{abs-1712-04851} has shown that it makes no difference in accuracy whether or not temporal order is reversed for the current state-of-the-art approach I3D. 

On the other hand, traditional datasets for video analysis such as UCF101~\cite{abs-1212-0402}, Sport1M~\cite{KarpathyTSLSF14}, and THUMOS~\cite{jiang2014thumos}, contain many actions that can be identified without temporal reasoning, which leads to approaches only good at modeling appearance invariance and short-term motion.

From the above points, we can see that most existing methods such as 3D CNNs~\cite{TranBFTP15} or two-stream networks~\cite{SimonyanZ14} are learned directly from raw pixels, which makes it difficult to recognize high-level actions from low-level trivial details efficiently.

In our opinion, a good approach for video understanding should encode the appearance and motion evolution in a certain temporal sequence effectively and efficiently. In this work, we seek to address two questions:
\begin{itemize}
\item How to encode spatial-temporal evolution in videos effectively and efficiently ?
\item Can we capture the spatial-temporal evolution correctly with the help of 2D CNNs?
\end{itemize}

Trying to answer the first question, our \textbf{first contribution} is to introduce a novel video representation called DenseImage which distills spatial-temporal evolution of a video into a matrix, where the spatial information is encoded in each column of the matrix and the temporal information is preserved in the row sequence.

To answer the second question, our \textbf{second contribution} is to propose a temporal-order-preserving CNN network to capture the core common spatial-temporal properties in DenseImage. The local temporal correlation constraint contained in the learning process makes the temporal evolution stable, and also captures temporal evolution at multiple time scales with different filter widths.

To sum up, a novel approach called DenseImage Network is proposed for video spatial-temporal encoding and understanding. Experiments on two challenging benchmarks and visualization analysis demonstrate that our DIN can accurately and efficiently capture the common spatial-temporal evolution between similar actions, with much less time-and-memory cost.

\section{Related work}
Early works for video understanding usually use hand-crafted features and SVM classifier, here we focus on recently proposed approaches based on convolutional neural networks.

\textbf{How to encode spatial-temporal evolution in videos effectively and efficiently ?} The success of static image classification with CNNs has driven the development of video recognition, but how to represent spatial-temporal evolution in videos using CNNs is still a problem. ~\cite{KarpathyTSLSF14} studied approaches for fusing information over
temporal dimension via 2D CNNs. ~\cite{SimonyanZ14} proposed a two-stream CNNs, one stream extract spatial information from RGB, and the other extract temporal information from optical flow, and finally the prediction scores from each stream are fused. ~\cite{WangXW0LTG16} proposed a temporal segment networks(TSN) for long-range temporal structure modeling, they extend the traditional two-stream method with a sparse temporal sampling strategy. ~\cite{TranBFTP15} proposed 3D CNNs to capture both appearance and motion features simultaneously. ~\cite{CarreiraZ17} found that optical flow is also useful in 3D CNNs, they took the strength of both two-stream and 3D CNNs achieved state-of-the-art performance on UCF101.

Recently, ~\cite{abs-1712-08416} pointed that the invariance to appearance is responsible for the success of optical flow, rather than the temporal trajectories.They shuffled flow fields, but the accuracy descrised slightly from 86.85\% to 78.64\%, they further shuffled the images to compute optical flow, and the accuracy is still upto 59.55\%. Their experiments illustrate that relay on optical flow to model temporal stucture is not enough yet. ~\cite{abs-1712-04851} has shown that it makes no difference in accuracy whether or not temporal order is reversed with the state-of-the-art approach I3D on Full-Kinetics dataset~\cite{CarreiraZ17}, one reason is that many kinds of actions in classical datasets can be identified by a single frame, which leads the existing approaches pay more attention to appearance and short-term motion. On the other hand, optical flow needs to be pre-computed before training which lowers the efficiency of the online recognition system and also requires lots of storage. Our DenseImage Network do not need optical flow to capture motion information, instead it further exploits the strength of 2D CNN to capture spatial-temporal evolution in video, which is data-efficient.

\textbf{Can we capture the spatial-temporal evolution accurately with the help of 2D CNNs?} As is known, CNNs perform well on the task of image classification, and it can also be used to other static image understanding tasks, such as: image caption, image semantic segmentation, however when it comes to video understanding, one may doubt that details are lost when using pretrained CNNs to distill spatial information in frame level, however, recently ~\cite{SantoroRBMPBL17} shows that relational network module using the features exracted from ImageNet pretrained CNNs such as ResNet101~\cite{HeZRS16}, can gain the ability of spatial relational reasoning and even outperforms average human performance on dataset CLEVR~\cite{JohnsonHMFZG17}, which is a visual question answering dataset designed to address relational reasoning in images. Inspried by that, ~\cite{abs-1711-08496} successfully extend relational networks to temporal relational reasoning by using multiple time scale MLPs to model temporal relation between frames. Their works demonstrate that pretrained CNN is still a powerful feature extractor for relational reasoning, which is a building block of our proposed video representation: DenseImage.

\begin{figure*}
	\begin{center}
		\includegraphics[width=1\linewidth]{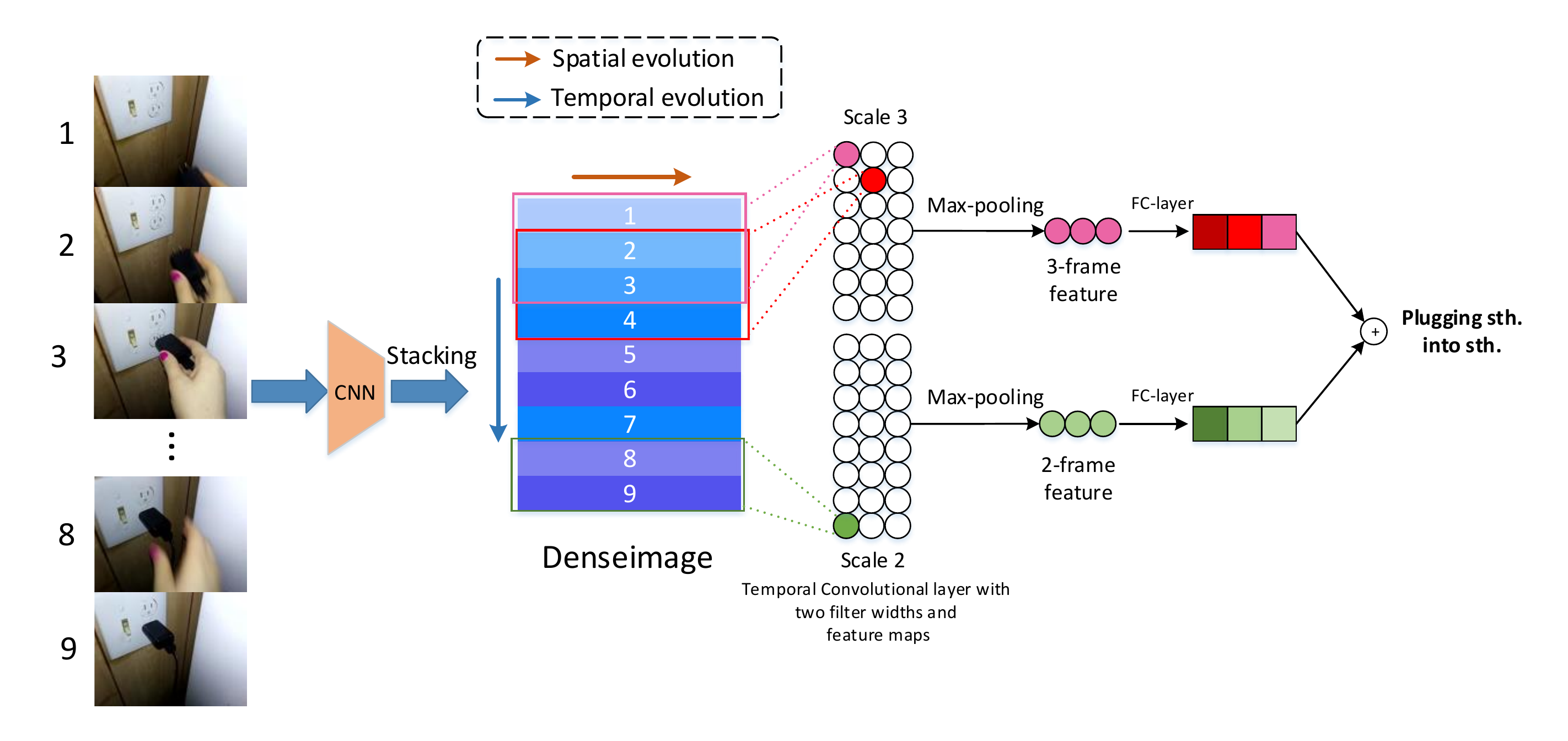}
		\caption{Framework of DenseImage Network, which consists of a video encoding method called DenseImage and a simple yet powerful learning strategy to capture the gist of spatial-temporal evolution. Here is an example with two different time scales.}
		\label{Figure 2.}
	\end{center}
	\vspace{-0.5em}
\end{figure*}
\textbf{The approaches similar to ours.} Dynamic image networks~\cite{BilenFGVG16} and TRN~\cite{abs-1711-08496} are the two most related works to ours. Dynamic image using rank pooling to compress a video into an image, it's good at modeling long term motion patterns, but loses details, while our method apply CNNs to distill spatial-temporal evolution of a video into a matrix called DenseImage. TRN apply MLPs to model temporal relational reasoning, however in order to calculate efficiently, they have to down sample from n-frame combinations in each time scale, in our opinion, this operation could cause unstableness of temporal evolution as the time interval in each scale between frames is randomly decided, not fixed, which makes their model converge slower, what's more, down sampling has the risk of losing frame which is important for action recognition. Instead, we propose a simple yet powerful temporal convolution based on DenseImage, it incorporates a local temporal correlation constraint and can effectively and efficiently capture temporal evolution at multiple time scales with different filter widths.

In brief, we propose an effective and efficient method which consists a novel video representation and a simple yet powerful learning strategy for video understanding tasks.

\section{Approach}
In this section,we give detailed descriptions of DenseImage Network module. As shown in Figure 2, it consists of a compact structure distilling the spatial-temporal evolution of a video into a matrix called DenseImage, and a temporal-order-preserving CNN network for video understanding.

\subsection{Video Encoding with DenseImage}
Suppose that we  sample n  frames from a video: \{$I_1,\cdots,I_n\}$, the form of encoding function $\psi$ is an open question, in this work we use ImageNet pretrained CNNs to form it. Let  $x_t=\psi(I_t) \in \mathbb{R}^k$ be a feature vector extracted from each individual frame $I_t$.  We then stack them in temporal order as below: $$X=x_1 \textcircled{+}x_2\textcircled{+} \cdots \textcircled{+}x_n. \eqno{(1)}$$ where $\textcircled{+}$ is the concatenation operator, the matrix $X \in \mathbb{R}^{nk}$ called DenseImage because each row in $X$ represents a frame. 

This encoding algorithm distills spatiotemporal evolution in a video into a DenseImage $X$, where the spatial information is encoded in each column of  $X$ and the temporal information is preserved in row sequence.
\subsection{Video Understanding with DenseImage Networks}
Given a DenseImage $X$ for a video, let $x_{i:i+j}$ refer to the concatenation of frames $x_i,x_{i+1},\cdots,x_{i+j}$. We then conduct convolution as below: $$c^h_{i,m}=f(w_{m,h}^T x_{i:i+h-1}+b_{m}). \eqno{(2)}$$ Where $m$ is the channel index of feature map, $w_{m,h}$ is the filter that captures temporal evolution between $h$ frames, $b_{m} \in \mathbb{R}$ is a bias term and $f$ is a non-linear function such as 
\begin{figure}
	\begin{center}
		\includegraphics[width=1\linewidth]{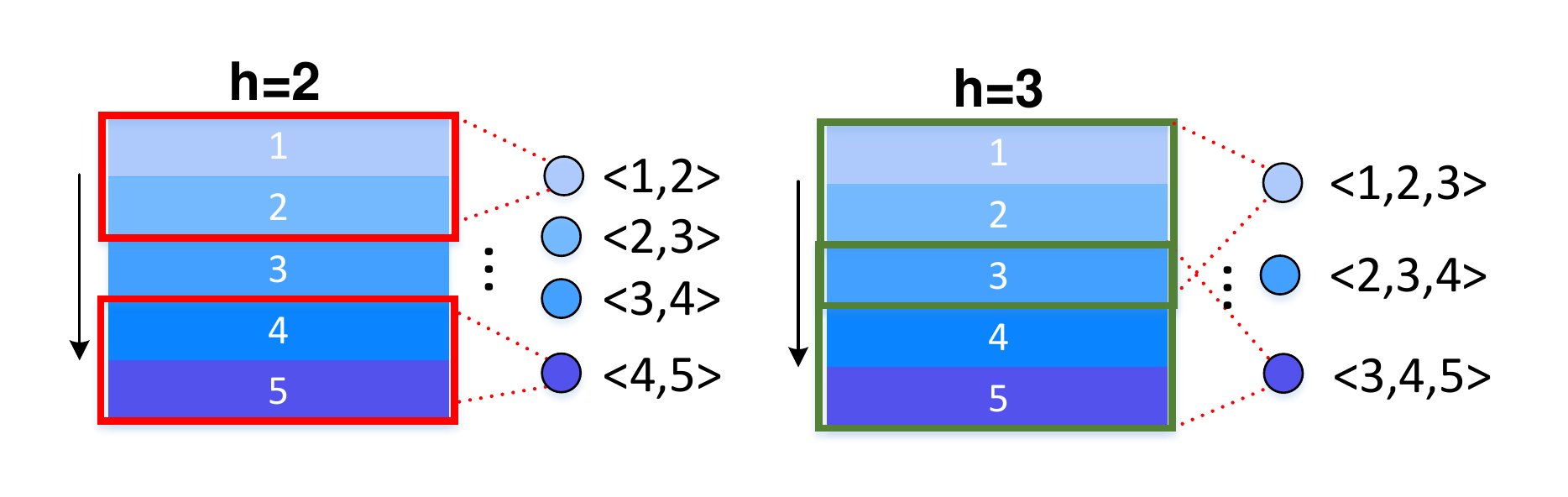}
		\caption{Spatial-temporal convolution on DenseImage with local temporal correlation constraint.}
		\label{Figure 2.}
	\end{center}
	\vspace{-0.5em}
\end{figure}
the rectifier. Filter $w_{m,h}$ is applied to each possible window of frames in $X$ to produce a feature map or a vector: $$\mathbf{c^h_m} =[c^h_{1,m},c^h_{2,m},\cdots,c^h_{n-h+1,m}], \eqno{(3)}$$ Note that each element in $\mathbf{c^h_m}$ represents a local temporal evolution in its position, and the whole vector $\mathbf{c^h_m}$ represents an abstract h-frame temporal evolution for $X$ with one channel $m$ and one filter size $h$. We then apply a max pooling operation , $o_m^h=max(\mathbf{c^h_m})$ ,  to get the most important local temproal structure.

Therefore, if the number of channels is $M$, then for filter size $h$, we'll get a h-frame spatial-temporal evolution representation based on DenseImage $X$, $$\mathbf{c^h} =[o_{1}^h,o_{2}^h,\cdots,o_{M}^h]. \eqno{(4)}$$

As shown in Figure 3, our approach is simple yet powerful at:

\begin{itemize}
\item \textbf{Local temporal correlation constraint.} Only adjacent $h$ frames will be convouted in filter $w_{m,h}$, this constrain makes temporal evolution much more stable, therefore the model is prone to be trained. 
\begin{figure*}[t]
	\begin{center}
		\includegraphics[width=1\linewidth]{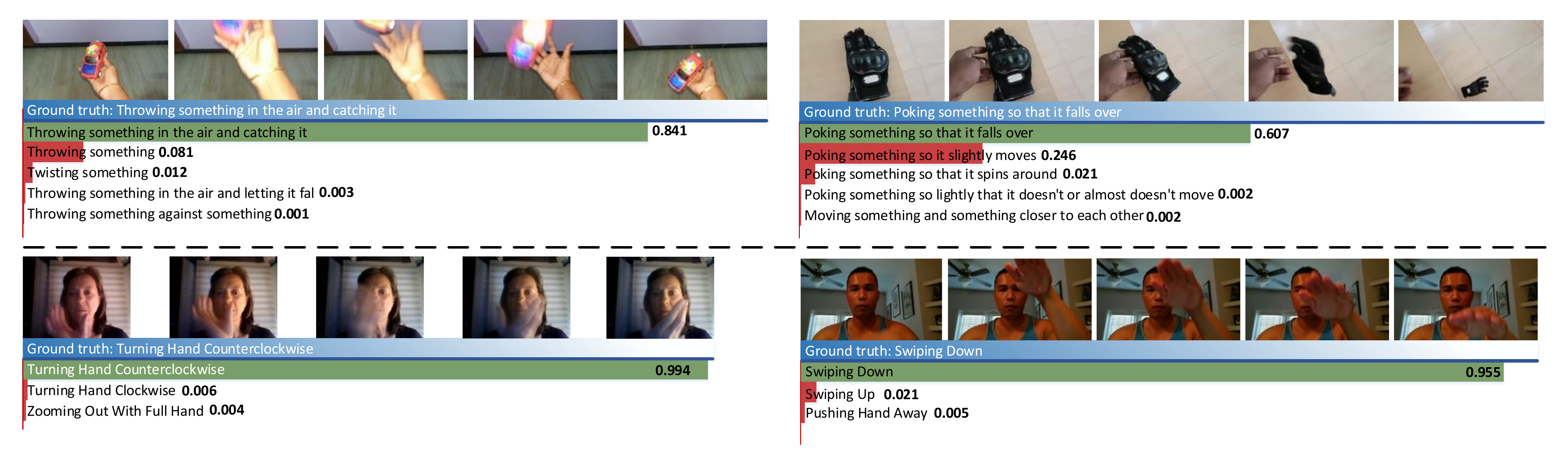}
		\caption{Action recognition results from two challenging new benchmarks Something-Something and Jester. Bars with different lengths indicating the recognition scores.}
		\label{Figure 4.}
	\end{center}
	\vspace{-1em}
\end{figure*}
\item \textbf{Temporal order preserving.} As shown in Figure 3, Suppose three ordered frames $[A,B,C]$ are sampled from a video, then a 2-frame filter works on it, getting a vector $\mathbf{c}=[c_1,c_2]$. Notice that $c_1$  represents temporal information for the ordered pair $<A,B>$ and $c_2$ for $<B,C>$.
\item \textbf{Efficient multi-scale temporal structure modeling.} Multi-scale temporal evolution can be captured by varying the value of $h$. For example, if $h=\{2,3,4\}$, it can model 2-frame, 3-frame, 4-frame temporal structure in videos.
\end{itemize}

Given multi-scale temporal features for a DenseImage, the classification can be formed as: $$score=softmax(\sum_{h\in H} f_{\phi(h)} (\mathbf{c^h})), \eqno(5)$$

Each timescale feature $\mathbf{c^h}$ is passed to a fully connected layer $f_{\phi(h)}$ with parameters $\phi(h)$, and the probability distributions are sumed and then normalized by a softmax function to get the final classification score. Note that our approach is differentiable throughout each module: DenseImage ecoding, convolution on DenseImage and classification, so they can all be trained together with back propagation algorithm.

\section{Experiments}
\subsection{Datasets}
We evaluate this work on two challenging new benchmarks, in which the spatial-temporal evolution between frames is critical to recognize actions correctly.

\textbf{Something-Something.}The dataset~\cite{GoyalKMMWKHFYMH17} is collected for generic human-object interaction. It comprises of 174 categories, like "Dropping something into something" and even "Pretending to open something without actually opening it". It contains 108,499 videos in total, 86,017 videos for training, 11,522 for validation and 10,960 for testing.
\begin{figure*}[t]
	\begin{center}
		\includegraphics[width=1\linewidth]{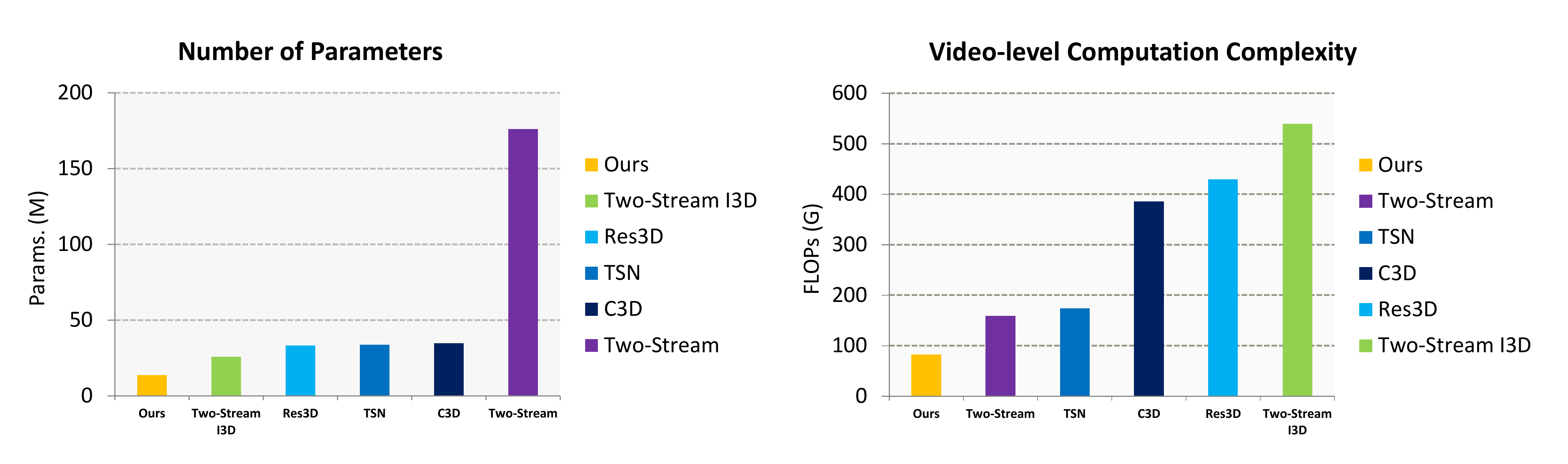}
		\caption{Number of parameters and video-level computation complexity of each method. Our method is 6.5x more efficient than the state-of-the-art two-stream I3D, and has much less parameters.}
		\label{Figure 5.}
	\end{center}
	\vspace{-0.5em}
\end{figure*}

\textbf{Jester.}The dataset~\cite{jester} is collected for gesture recognition. It comprises of 27 categories, like "Swiping Left" and "Pulling Two Fingers In". It contains 148,092 videos in total, 118,562 videos for training, 14,787 for validation and 14,743 for testing.Each video is represented as a set of images that are extracted from the orginal videos at 12 frames per second for both datasets.

\subsection{Implementation details}

\textbf{Features:}As it known that ImageNet pretrained CNNs are powerful for image representation and superior CNNs such as ResNet~\cite{HeZRS16} usually perform better. Here in order to verify the effectiveness of our method, we fix the ImageNet pretrained CNNs to be the same in this work. We adopt Inception with Batch Normalization(BN-Inception)~\cite{IoffeS15} because of its balance between accuracy and efficiency. Specifically, image features are extracted from the $global\_pool$ layer with dimension 1024, following a fully connected layer to reduce dimension from 1024 to 256.

\textbf{DenseImage:}Following TSN~\cite{WangXW0LTG16}, we apply the same sampling strategy to sample eight frames from each video. Therefore DenseImage $X\in \mathbb{R}^{8*256}$ (Section 3.1 for details).

\textbf{Training settings:}We empirically set the timescale $h=\{2,3,\cdots,6\}$, filter numbers of each time scale is 256. We follow the strategy of partial BN and dropout after global pooling as used in TSN. We add an extra dropout layer before fully connected layer to further reduce the effect of over-fitting. Mini-batch SGD algorithm is applied to optimize our model. We use mini-batches of 32 videos, momentum of 0.9 and weight decay of $5e^{-4}$. All models are initialized with learning rate $5e^{-4}$ and this value is further reduced to its $\frac{1}{10}$ whenever the validation error stops decreasing.

\begin{table}[t]
\centering
\renewcommand\arraystretch{1.2}
\caption{Top-1 accuracy on Something-Something test set.}
\begin{tabular}{p{5cm} p{2cm}<{\centering}}
\toprule[1.4pt]
Model&Top 1 acc.(\%)\\
\hline
Peter\_CV&19.68\\
Valentin(esc)&24.55\\
Harrison.AI&26.38\\
I3D&27.23\\
Besnet&31.66\\
TRN\_v2&33.60\\
\hline
DIN(ours)&\textbf{34.11}\\

\bottomrule[1.5pt]

\end{tabular}
\end{table}

\begin{table}[t]
\centering
\renewcommand\arraystretch{1.2}
\caption{Top-1 accuracy on Jester test set.}
\begin{tabular}{p{5cm} p{2cm}<{\centering}}
\toprule[1.4pt]
Model&Top 1 acc.(\%)\\
\hline
3D CNN&77.85\\
20BN's Jester System&82.34\\
ConvLSTM&82.76\\
VideoLSTM&85.86\\
Ford's Gesture Recognition System&94.11\\
Besnet&94.23\\
TRN&94.78\\
\hline
DIN(ours)&\textbf{95.31}\\

\bottomrule[1.5pt]

\end{tabular}
\end{table}

\subsection{Accuracy}
We show the leaderboard on Something-Something\footnote{https://www.twentybn.com/datasets/something-something} and Jester\footnote{https://www.twentybn.com/datasets/jester} datasets. As shown in Table 1 and Table 2, we tops the leaderboard of Something-Something and Jester at the time of submission. Notice that we only use single modal RGB features and without ensembling of multiple models.

For intuitive explanation, we show four examples in Figure 4. As can be seen, one can hardly identify the four examples with only one single frame, because spatial-temporal evolution in them is essential for a successful recognition. Our DIN method correctly identifies these examples indicating that it captures the spatiotemporal evolution between frames. It further demonstrate that the cooperation of proposed video representation:DenseImage and temporal convolution is effective to recognize these actions.

\subsection{Efficiency Analysis}
Our method is efficient because there is no extra data, such as optical flow, needed to be pre-computed and the temporal convolution based on DenseImage is an efficient 2D style CNN which contains only one single convolutional layer. 

Figure 5 compares the number of parameters and the video-level computation complexity of our approach with state-of-the-art method: Two-Stream~\cite{SimonyanZ14}, TSN~\cite{WangXW0LTG16}, C3D~\cite{TranBFTP15}, Res3D~\cite{abs-1708-05038}, Two-Stream I3D~\cite{CarreiraZ17}. As can be seen, our approach has 1.9x much less parameters than the current state-of-the-art Two-Stream I3D, and 12.9x less for the classical Two-Stream, TSN use BN-Inception as base model which is the same with us, however due to using 3 modalities: RGB, optical flow and warped flow it has to fine-tune 3 base model to capture feature in different modalities which leads 3x much more parameters than us. 

As for computation complexity, state-of-the-art methods are usually applied to multiple video clips and the recognition results are averaged during test time, for fair comparison here we compare the video-level computation complexity of them. Our approach is 6.5x efficient than Two-Stream I3D, this is mainly due to Two-Stream I3D is trained on 4x longer videos , and the optical flow stream further increases the computation complexity. 

To sum up, the optical flow computation is the bottleneck for two-stream networks, and 3D CNNs are computationally expensive based on dense sampled frames. In contrast, there is only RGB discrete frames used in our approach, and all of the convolutional layers in our DIN is 2D, which makes DIN efficient.

\subsection{Visualization Analysis}
This section we conduct several experiments to visualize the spatial-temporal evolution captured by our DIN.

\begin{figure*}[!htbp]
	\begin{center}
		\includegraphics[width=1\linewidth]{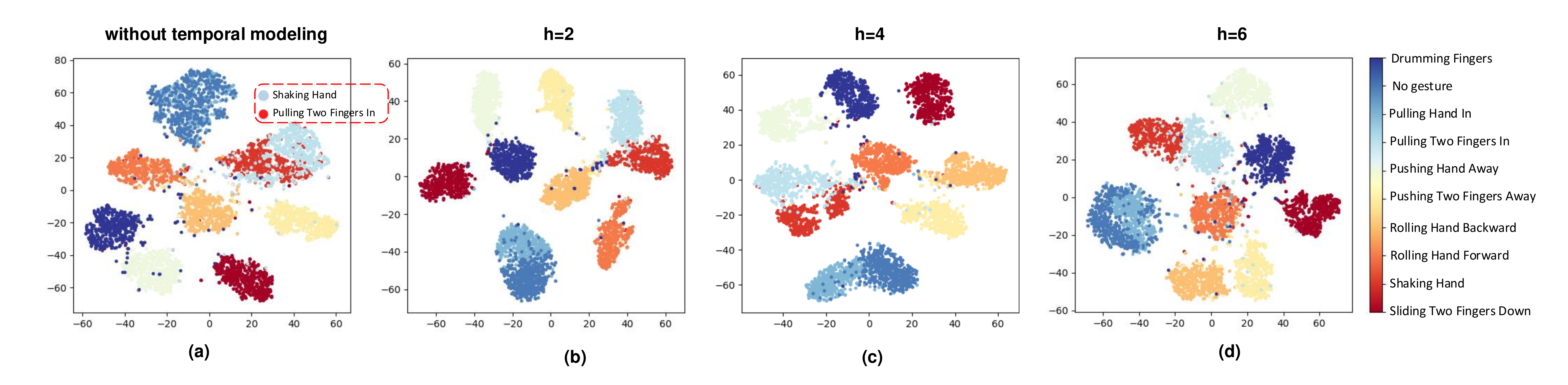}
		\caption{t-SNE plot showing that complex actions can be better classified at different time scales. As can be seen in (a), "Shaking Hand" is highly overlapped with "Pulling Two Fingers In", indicating that temporal evolution is essential for a successful recognition. As shown in (b) and (c), samples from different classes are clearly separated, indicating that filter widths with 2 and 4 can capture the tiny difference between similar classes. As shown in (d), "No gesture" and "Pulling Hand In" is clustered together, while other classes such as:"Rolling Hand Backward" and "Rolling Hand Forward" are more distinguishable, indicating that the cooperation  of different time scales is important to recognize these samples.}
		\label{Figure 6.}
	\end{center}
	\vspace{-0.5em}
\end{figure*}
\begin{figure*}[h]
	\begin{center}
		\includegraphics[width=1\linewidth]{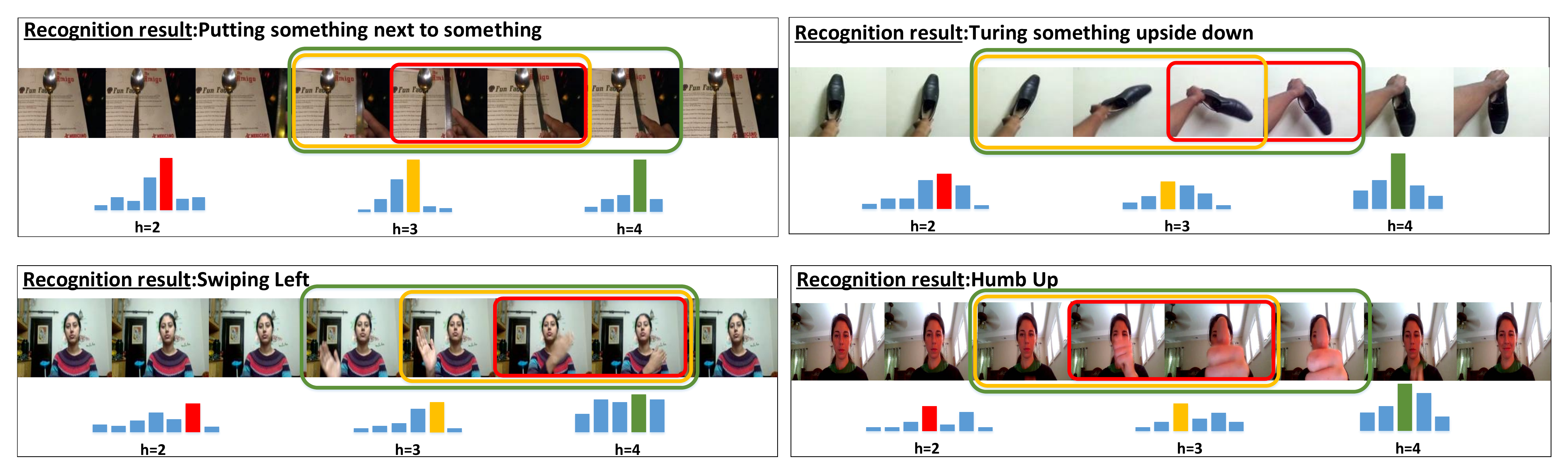}
		\caption{Visualization the response of filters with different widths in videos. Showing that the common spatial-temporal evolution in videos can be correctly captured at different time scales. The height of each bin represents the response intensity at corresponding position, h refers to the width of filter, and the bounding boxs in different colors correspond to positions where the response is greatest. Eight frames sampled from each video, so there exists 7 bins with $h=2$, 6 bins with $h=3$, and 5 bins with $h=4$.}
		\label{Figure 7.}
	\end{center}
	\vspace{-0.5em}
\end{figure*}

\textbf{Temporal convolution based on DenseImage can accurately capture the common spatial-temporal evolution between similar actions at multiple time scales.} As shown in Figure 6, we use t-SNE algorithm to visualize the high-level video features with different filter widths, features without temporal modeling are  also included. These videos come from the 10 most frequent action classes in the Jester validation set.

As described in section3, our DIN consists a 2D temporal convolutional layer which can capture spatial-temporal evolution at multiple time scales with different filter widths. Here we visualize the response of different width of filters at ever possible location in DenseImage. As shown in Figure 7, which indicating that our DIN can discover frames that are important for action recognition at multiple time scales.

\section{Conclusion}
In this work, a novel approach called DenseImage Network is proposed for video spatial-temporal encoding and understanding. Experiments on two challenging benchmarks and visualization analysis demonstrate that our DIN can accurately and efficiently capture the common spatial-temporal evolution between similar actions, with much less time-and-memory cost.

\bibliographystyle{named}
\bibliography{ijcai18}

\end{document}